# Kannada Character Recognition System: A Review

[1]K. Indira, [2]S. Sethu Selvi
[1]Assistant Professor, indm_1@yahoo.com
[2]Professor and Head, selvi_selvan@yahoo.com
[1,2]Department of Electronics and Communication Engineering
M. S. Ramaiah Institute of Technology, Bangalore – 560 054

**Abstract**

**Intensive research has been done on optical character recognition ocr and a large number of articles have been published on this topic during the last few decades. Many commercial OCR systems are now available in the market, but most of these systems work for Roman, Chinese, Japanese and Arabic characters. There are no sufficient number of works on Indian language character recognition especially Kannada script among 12 major scripts in India. This paper presents a review of existing work on printed Kannada script and their results. The characteristics of Kannada script and Kannada Character Recognition System kcr are discussed in detail. Finally fusion at the classifier level is proposed to increase the recognition accuracy.**

## 1. INTRODUCTION

Kannada, the official language of the south Indian state of Karnataka, is spoken by about 48 million people. The Kannada alphabets were developed from the Kadamba and Chalaukya scripts, descendents of Brahmi which were used between the $5^{th}$ and $7^{th}$ centuries A.D. The basic structure of Kannada script is distinctly different from the Roman script. Unlike many north Indian languages, Kannada characters do not have shirorekha (a line that connects all the characters of any word) and hence all the characters in a word are isolated. This creates a difficulty in word segmentation. Kannada script is more complicated than English due to the presence of compound characters. However, the concept of upper/lower case characters is absent in this script.

Modern Kannada has 51 base characters, called as Varnamale. There are 16 vowels and 35 consonants. (Table 1). Consonants take modified shapes when added with vowels. When a consonant character is used alone, it results in a dead consonant (mula vyanjana). Vowel modifiers can appear to the right, top or at the bottom of the base consonant. Table II shows a consonant modified by all the 16 vowels. Such consonant-vowel combinations are called live consonant (gunithakshra). When two or more consonant conjuncts appear in the input they make a consonant conjunct (Table III). The first consonant takes the full form and the following consonant becomes half consonant. In addition, two, three or four characters can generate a new complex shape called a compound character.

The number of possible Consonant–Vowel combinations is $35 \times 16 = 560$. The number of consonant -consonant vowel combinations is $35 \times 35 \times 16 = 19600$. If each akshara is considered as a separate category to be recognized, building a classifier to handle these classes is difficult. Most of the aksharas are similar and differ only in additional strokes, it is feasible to break the aksharas into their constituents and recognize these constituents independently. The block diagram of a KCR System is shown in Figure 1.

## 2. PRE PROCESSING

In a KCR system, the sequences of preprocessing steps are as follows:

### 2.1 Binarization

The page of text is scanned through a flat bed scanner at 300dpi resolution and binarized using a global threshold computed automatically based on a specific image [1]. A binary image is obtained by considering the character as ON pixels and the background as OFF pixels. The binarized image is processed to remove any skew so that text lines are aligned horizontally in the image.

### 2.2 Skew Detection

Methods to detect and correct the skew for the document containing Kannada script has not been reported in any of the recognition methods for Kannada script. However, the methods discussed here briefly work on the document containing Kannada script.

#### 2.2.1 Projection profiles

A projection profile is a histogram of the number of ON pixels accumulated along parallel sample lines taken through the document. The profile may be at any angle, but often it is taken horizontally along rows or vertically along columns, and these are called the horizontal and vertical projection profiles respectively. For a document whose text lines span horizontally, the horizontal projection profile will have peaks whose widths are equal to the character height and minimum height valleys whose widths are equal to the between line spacing. The most straight forward use of the projection profile for skew detection is to compute it at a number of angles close to the expected orientations [2]. For each angle, a measure is made of the variation in the bin heights along the profile, and the one with the maximum variation gives the skew angle. At the correct skew angle, since scan lines are aligned to text lines the projection profile has maximum height peaks for text and valleys between line spacing. Some of the Indian scripts, such as Devanagiri, have a dark top line linking all the characters in a word. This strong linear feature can be used for projection profile based method. In Kannada, such a line linking all characters of the word is not present. However, a short horizontal line can usually be seen on top of most of the characters. This feature can be exploited for skew estimation.

#### 2.2.2 Hough Transform

This transform [3] maps points from *(x, y)* domain to curves in *(ρ, θ)* domain, where $\rho$ is the perpendicular distance of the line from the origin and $\theta$ is the angle between the perpendicular line and horizontal axis in the *(x, y)* plane. Crossing curves in *(ρ, θ)* domain is the result of collinear pixels in *(x, y)* plane. $\theta$ is varied between 0 and $180^0$ for each black pixel in a document image and $\rho$ is calculated in Hough space. The maximum value in *(ρ, θ)* space is considered as the skew angle of the document image. This method is time consuming due to the mapping operation from *(x, y)* plane to *(ρ, θ)* plane for all black pixels, especially for images containing non text dominant area (i.e. pictures, graphs etc.). Le et al [4] improved the computational time by applying Hough transform only on the bottom pixels of connected components belonging to the dominant text area. This method gives an accuracy of $0.5^0$ to the original skew angle of the image. Hough transform based methods are robust enough to estimate skew angles between $-15^0$ to $15^0$. But they are computationally expensive and sensitive to noise.

#### 2.2.3 Wavelet decomposition and projection profiles

Skewed document images are decomposed by the wavelet transform [5]. The matrix containing the absolute values of horizontal subband coefficients which preserves the text horizontal structure, is then rotated through a range of angles. A projection profile is computed at each angle, and the angle that maximizes a criterion function is regarded as the skew angle. This algorithm performs well on document images of various layouts and is robust to different languages and hence can be used for documents containing Kannada script.

#### 2.2.4 Wavelet decomposition and Hough Transform [3, 5]

The document image is decomposed by wavelet transform and the LL subband which preserves the original image is rotated through a range of angles and Hough Transform is calculated at each angle. The angle that maximizes the highest number of counts corresponds to the skew angle of the text. This method is suitable for Kannada text scanned at 300dpi and is faster compared to other methods.

### 2.3 Skew Correction

Skew correction is performed by rotating the document through an angle $-\theta$ with respect to the horizontal line, where the detected angle of skew is $\theta$. In order to prevent the image being rotated off the image plane, the skewed image is first translated to the center and the new image dimensions are computed.

## 3. SEGMENTATION

Segmentation is the process of extracting objects of interest from an image. The first step in segmentation is detecting lines. The subsequent steps are detecting the words in each line and the individual characters in each word.

### 3.1 Line Segmentation

Kunte and Samuel describe horizontal and vertical projection profiles (HPP and VPP) [7] for line and word detection respectively. The horizontal projection profile is the histogram of the number of ON pixels along every row of the image. White space between text lines is used to segment the text lines. Figure 2 shows a sample Kannada document along with its HPP. The projection profile has valleys of zero height between the text lines. Line segmentation is done at these points.

Kumar and Ramakrishnan have reported segmentation techniques for Kannada script. The bottom conjuncts of a line overlap with top matras [8] of the following text lines in the projection profile. This results in nonzero valleys in the HPP.

These lines are called Kerned Text lines. To segment such lines, the statistics of the height of the lines are found from the HPP. Then the threshold is fixed at 1.6 times the average line height. This threshold is chosen based on experimentation of segmentation on a large number of Kannada documents. Nonzero valleys below the threshold indicate the locations of the text line and those above the threshold correspond to the location of kerned text lines. The midpoint of a nonzero valley of a kerned text line is the separator of the line. Ashwin and Shastry have solved the problem of overlapping of consonant conjunct of one line with the vowel modifier of the next line by extracting the minima of the horizontal projection profile smoothed by a simple moving average filter. The line breaks obtained sometimes segment inside a line. Such false breaks are removed by using statistics of line widths and the separation between lines.

### 3.2 Word Segmentation

Kunte and Samuel have proposed word segmentation by taking the vertical projection profile of an input text line. For Kannada script, spacing between the words is greater than the spacing between characters in a word. The spacing between the words is found by taking the vertical projection profile of an input text line. The width of zero valued valleys is more between the words in line as compared to the width of zero valued valleys that exist between characters in a word. This information is used to separate words from the input text lines. Figure 3 shows a text line with its VPP.

Kumar and Ramakrishnan have described that Kannada words do not have shirorekha and all the characters in a word are isolated. Further, the character spacing is non-uniform due to the presence of consonant conjuncts. Thus, spacing between the base characters in the middle zone becomes comparable to the word spacing. This could affect the accuracy of word segmentation. Hence, morphological dilation is used to connect all the characters in a word before performing word segmentation. Each ON pixel in the original image is dilated with a structuring element then VPP of the dilated image is determined. The zero valued valleys in the profile of the dilated image separates the words in the original image. Figure 4 illustrates the dilated image and VPP of a line. The accuracy of word segmentation depends upon the structuring element and the type of structuring element has to be decided by performing experiments on different text with various fonts. Ashwin and Shastry have suggested to adapt a threshold for each line of text to separate interword gaps from inter-character gaps. Threshold has to be obtained by analyzing the histogram of the width of the gaps in a line. Hence, the threshold is not fixed and it has to be changed after performing the histogram of a line. This method may not work for all types of document.

### 3.3 Character Segmentation

#### 3.3.1 Three Stage Character Segmentation

Kumar and Ramakrishnan [8] described a three-stage character segmentation for separating Kannada characters from the segmented word. Three line segmentation of character involves the division of each into three segments: Top zone consists of top matras, middle zone consists of base and compound characters and bottom zone consists of consonant conjuncts. Head line and base line information is extracted from the Horizontal Projection Profile. Head line refers to the index corresponding to maximum in the top half of the profile base line refers to the index corresponding to maximum in the bottom half of the profile. Using the baseline information, text region in the middle and top zones of a word is extracted and its VPP is obtained. Zero valued valleys of this profile are the separators for the characters as in Figure 5. Sometimes, the part of a consonant conjunct in the middle zone is segmented as a separate symbol then, split the segmented character into a base character and a vowel modifier (top or right matra). The consonant conjuncts are segmented separately based the on connected component analysis (CCA).

#### 3.3.2.1 Consonant Conjunct Segmentation

Knowledge based approach is used to separate the consonant conjuncts. The spacing to the next character in the middle zone is more for characters having consonant conjuncts than for the others. To detect the presence of conjuncts, a block of partial image in the bottom zone corresponding to the gap between adjacent characters in the middle zone is considered. If the number of ON pixels in the partial image exceeds a threshold (for examples 15 pixels), a consonant conjunct is detected. Sometimes, a part of the conjunct enters the middle zone between the adjacent characters such parts will be lost if the conjunct is segmented only in the bottom zone. Thus, in order to extract the entire conjunct CCA is used. However, in some cases the conjunct is connected to the character in the middle zone causing difficulty in using CCA for segmenting the conjunct alone. This problem is solved by removing the character in the middle zone before applying CCA. But the consonant conjunct segmentation described in this paper will have discrepancies for some of the words as depicted in Figure 6. The normal segmentation technique described in this paper would falsely recognize the consonant conjunct for both the characters. This discrepancy can be solved by isolating the bottom matra from the rest of the characters. The height and width of the consonant conjunct is determined. The position of the extreme right ON pixel is marked and the next character is scanned from the next column. By doing this, the consonant conjunct is recognized by the first character and not with the second.

#### 3.3.2.2 Vowel Modifier Segmentation

This includes segmentation of the top and right matras. The part of the character above the headline in the top zone, is the top matra. Since the headline and baseline of each character is known, if the aspect ratio of the segmented character in the combined top and middle zone is more than 0.95, then it is checked for the presence of the right matra.

For right matra segmentation, three subimages of the character is considered: whole character, head and tail images. The head image is the segment containing five rows of pixels starting from the head line downwards similarly, the tail image contains five rows downwards from the baseline. The VPP for each of these images are determined. The index corresponding to the maximum profile of the character image is determined say *(P)*. The indices corresponding to the first zero values, immediately after the index *(P)* in the profiles of head and tail images, say *b1* and *b2* respectively, are determined. The break point is selected as the smaller of *b1* and *b2*.

### 3.3.2 Two Stage Character Segmentation

Kunte and Samuel have proposed a two stage method for segmentation of Kannada characters. As Kannada is a non-cursive script, the individual characters in a word are isolated. Spacing between the characters can be used for segmentation. But sometimes in VPP of a word, there will be no zero valued valleys, due to the presence of conjunct-consonant (subscripts) characters. The subscript character position overlaps with the two adjacent main characters in vertical direction.

In these cases the usual method of vertical projection profile to separate characters is not possible. In these cases the following two stage [7] approach is used,

Stage 1:

a. Check for the presence of subscripts in a word.
b. If subscripts are present, they are extracted first from the word using Connected Component method.

Stage 2:

a. Remaining characters from the word are extracted using VPP.
b. If subscripts are not present in a word then the characters from the word are extracted using VPP in one stage itself.

Thus for character segmentation it is first necessary to check whether there are any subscripts in a word. For this, a Kannada word is divided into different horizontal zones as described in Section 3.3.2.1

#### 3.3.2.1 Zones in a Kannada word

A Kannada word can be divided into different horizontal zones. Two different cases are considered, a word without subscripts as in Figure 7 and a word with subscripts as in Figure 8. Consider the sample word as in Figure 7 which does not have a subscript character. The imaginary horizontal line that passes through the top most pixel of the word is the top line. Similarly, the horizontal line that is passing through the bottom most pixel of the main character is the base line. The horizontal line passing through the first peak in the profile is the head line. The word can be divided into top and middle zones. Top zone is the portion between the top and head line and the middle zone is the portion between the head line and base line.

For words with conjunct-consonant characters, it is divided into three horizontal zones as in Figure 8 for a sample word with subscripts. The word is divided into top, middle and bottom zones. The top and middle zones are chosen similar to that of the word without subscripts. A bottom portion is chosen between the baseline and the bottom line. The bottom line is the horizontal line passing through the bottom most pixel of the word.

Before character segmentation it is first necessary to find out whether the segmented word has a subscript or not. This can be detected as follows:

i. In the horizontal projection profile as in Figure 7, there are two peaks of approximately equal size in the top and middle zones of the word. The absence of the third peak after the second peak indicates that there are no subscripts in the word.
ii. In the HPP as in Figure 8, there are two peaks of approximately equal size in the top and middle zones of the word. Also, there is an occurrence of third peak after the second peak in the bottom zone of the word, which is due to the subscripts in the word.

Thus, by checking the presence or absence of the third peak in the bottom zone of the horizontal projection profile of the segmented Kannada word, it is possible to find out whether the segmented word has a subscript or not.

#### 3.3.2.2 Character Segmentation of a word without Subscripts

Consider a Kannada word which does not have any subscripts. There is zero valued valleys in the VPP of the word which makes the character separation easier. The portion of the image which lies between two successive zero valued valleys of the VPP is assumed to be as a separate character and separated out.

#### 3.3.2.3 Character Segmentation of a word having subscripts

Consider a sample Kannada word as in Figure 8, which contains subscripts. If VPP of this word is considered, then there will be no zero valued valleys between the first character 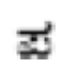 , its subscript character 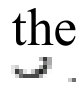 and also for the third character and its subscript. Hence, just the zero valued valleys of the vertical projection do not determine the character separation. The individual characters in this case are separated in two stages as follows:

In the first stage, the subscripts of the word are separated. In the second stage, from the plain word the individual characters are extracted using VPP.

**Stage 1 Subscript Character Segmentation**

Consider a sample word as shown in Figure 8. The total height of the word in terms of number of rows *(H)* is calculated. The columns of the word are scanned from left to right. Every column is scanned from bottom to top to find the presence of an ON pixel *P*. When such an ON pixel is found, the number of rows that has gone up *(L)* is counted. If *L* is less than or equal to some threshold value, the pixel *P* is assumed to be one of the points of the subscript character. Then using *P* as initial point, connected component algorithm [9] is applied to extract the subscript character at that position. Threshold value is calculated by finding the position of the valley between the second peak and the third peak which is below the base line in the bottom zone of the word. The scanning process is repeated till the end of the word (right most column) to extract all the subscript characters present in the word. At the end of Stage 1, after separating subscripts what remains is a plain word without having any subscript characters as in Figure 7.

**Stage 2 Main character segmentation:**

The output of the first stage converts the word with subscripts into a plain word without any subscript characters. Hence during the second stage, the same method used for segmenting the characters (for a word without subscripts) is followed for segmenting the main characters.

### 3.3.3. Over Segmentation and Merge approach

Ashwin and Sastry [1] have described the segmentation and merge approach for segmentation of a Kannada word. In this method the words are vertically segmented into three zones. This segmentation is achieved by analyzing the HPP of a word. Separating the middle zone from the bottom zone is easier as the consonant conjuncts are disconnected from the base consonant. Separating the top zone from the middle zone is difficult and there are situations where the top zone may contain some of the consonant or the middle zone may contain a little bit of the top vowel modifier. These inaccuracies are taken care by training the pattern classifier. Then the three zones are segmented horizontally. The middle zone is first over segmented by extracting points in the vertical projection showing valleys in the histogram exceeding a fixed threshold. The threshold is kept low so that a large number of segments are obtained. This segmentation does not give consistent segments, these segments are merged using heuristic merging algorithm and recognition based algorithm.

In the three stage segmentation technique proposed in [8], a character is decomposed into base character, vowel modifier and consonant conjuncts. Features are extracted from individual constituents of a character and each of the constituents are classified and then merged to form a class label. With this segmentation approach, the design of the classifier is complex. But the number of classes required to classify a character is *102* (vowels *(16)*, consonants *(35)*, vowel modifier *(16)*, consonant conjuncts *(35)*, $16+35+16+35 = 102$).

In the two stage segmentation procedure proposed in [7], the segmentation algorithm is simple and the design of the classifier is also simple as there is no need to merge the segmented symbols to form a class label. With this approach the number of classes required to classify a character is *646* (vowels *(16)*, consonants *(35)*, consonant vowel modifier *(35 x 16 = 560)*, and consonant conjuncts *(35))*.

In the over-segmentation and merge approach

[1], the segmentation algorithm does not give consistent segments and the segments may differ depending on the font and size of the characters and also gives a large number of small segments and cannot be merged using the merging algorithm.

## 4. FEATURE EXTRACTION

Features are a set of numbers that capture the salient characteristics of the segmented image. As font and size independent recognition is required, template matching is not advisable for recognition of segments. Segmented symbols are unequal in size and they are normalized to one size so that the classifier is immune to size changes in the characters. There are different features proposed for character recognition [10]. In spatial domain method Hu's invariant moments and Zernike moments [11] are used to represent the segmented Kannada character. In frequency domain methods [8] Discrete Cosine Transform (DCT), Discrete Wavelet Transform (DWT) and Karhunen Louve Transform (KLT) are used as features for recognizing Kannada characters. Ashwin and Sastry [1] have used a set of features by splitting each segment image into a number of zones. Kumar and Ramakrishnan describe a subspace projection for base characters. This subspace projection represents higher dimensional data in lower dimensional space by projecting them onto the subspace spanned by the eigen vectors corresponding to significant eigen values of the covariance matrix. Since characters in Kannada have a rounded appearance, the distribution of pixels in the radial and angular directions is considered. Table IV summarizes the various feature representations for Kannada characters.

## 5. CHARACTER CLASSIFICATION

The feature vector extracted from the segmented and normalized character has to be assigned a label using a character classifier. Methods for designing a character classifier are Bayes classifier based on density estimation, nearest neighbour classifier based on a prototype, linear discriminant functions and neural networks. The data set is divided into training set and test set for each character.

**Nearest Neighbour Classifier**:

The Euclidean distance [13] of the test pattern to all the vectors in the training pattern is computed and the test pattern is assigned to the class of the sample that has the minimum Euclidean Distance.

**K – Nearest Neighbour Classification** [13]**:**

Given a set of prototype vectors,

$$T_{XY} = \{(x_1, y_1), (x_2, y_2), \dots (x_l, y_l)\},$$

The input vectors being $x_i$ $X$ $R$ and corresponding targets being $y_i$ $Y$ $\{1,2,\dots c\}$ Let $R^n$ $(x)$ $\{x' : || x x' || r^2\}$ be a ball centered in the vector $x$ in which $K$ prototype vectors $x_i$ , $i$ $\{1,2,\dots K\}$ lie ie. $| x_i : x_i R^n (x) | K$ . The K nearest neighbour classification rule q: $X$ $Y$ is define as $q(x)$ $arg\ max\ v(x, y)$ , where $v(x,y)$ is the number of prototype vectors $x_i$ with targets $y_i$ $=y$, which lie in the ball $x_i$ $R$ $(x)$

**Back Propagation Network [BPN]:** This network is a multilayer perceptron with input layer, one or more hidden layer and output layer. BPN [8] is trained in batch mode using supervised learning, employing log sigmoidal activation function. The input is normalized to a range of 0 to 1 to meet the requirements of the activation function before training.

**Radial Basis function Network (RBF):** This is a three layer network consisting of input, hidden and output layers. The radial basis functions are centered on each training pattern and the layer biases are all kept constant depending on the spread of the Gaussian.

**Support Vector Machine (SVM):** The SVM classifier is a two class classifier based on the discriminant functions. A discriminant function represents a surface, which separates the patterns as two classes. For OCR applications a number of two class classifiers are trained with each one distinguishing one class from the other. Each class label has an associated SVM and a test example is assigned to the label of the class whose SVM gives the largest positive output. The example is rejected if no SVM gives a positive output.

Table V summarises the nearest neighbour classifier on various features. This classifier performance is 92.86% for spatial domain features and 98.83% for frequency domain features using DWT (Haar) for the base character. Thus the recognition rate of a nearest neighbour classifier for frequency domain features is higher compared to spatial domain features and is true for vowel modifier and consonant conjuncts. Table VI summarizes the neural network classifier on various features. Performance of BPN with frequency domain features is 95.07 % on an average for base character. BPN network is not used for spatial domain features. Thus, the performance of nearest neighbour classifier is better than BPN network. RBF [8] with Zernike moments, the recognition rate is 96.8%. With DWT and with structural features the recognition rate is 99.1% for the base character. RBF is used for spatial domain and frequency domain features. Thus performance of RBF network is higher than NN and BPN classifier and is true for vowel modifier and consonant conjuncts. SVM [1] with Zernike features, the recognition rate is 92.6%. With modified structural features, the recognition rate is 93.3% for the base character. Performance of SVM with spatial domain features is higher than NN classifier for the base character. Comparing the work proposed in [1], [8] and [14] RBF gives highest recognition rate of 99% using Haar features.

## 6. CONCLUSION AND FUTURE DIRECTION

In this paper, a complete character recognition system for printed Kannada documents is discussed in detail. Different segmentation techniques and various classifiers with different features are also discussed. Different segmentation methods lead to different classifier design, as it depends on the number of classes at the output stage of a classifier.

Results presented in [1] are based on scanning Kannada texts from magazines and textbooks at 300 dpi. The training and test patterns were generated from the same text and was ensured that these patterns are disjoint. SVM classifier is used in the recognition stage, and this type of two class classifier becomes complicated for a multi-class problem. This multi-class classification problem can be solved [16] by using pair wise based SVM classifier, fuzzy pair wise SVM and directed acyclic graph (DAG).

To improve recognition performance, a recognition based segmentation method uses dynamic wide length to provide segmentation points which are confirmed by the recognition stage. In [8] and [14] DWT, DCT, KLT, Hu's invariant moments and Zernike moments are used to represent each character and is classified using NN, BPN and RBF network. Results presented in [8] are based on presegmented characters and hence a complete system for recognition of Kannada text is not designed.

In [6] and [8], to increase the recognition accuracy structural features are included for disambiguating confused characters. Results presented in [11] and [14] are based only on base characters and consonant conjuncts. A complete KCR system is not proposed.

Curvelets give very good representation of edges in an image, it has high directional sensitivity and are highly anisotropic. Hence, can be used to represent Kannada characters for classification. Different types of features can be used to represent a character. Each of the feature vector is analysed using a classifier. Multiple categories [17] of the features are combined into a single feature vector and a final classifier provides the decision of the class of the test character.

Table 1. Vowels and Consonant

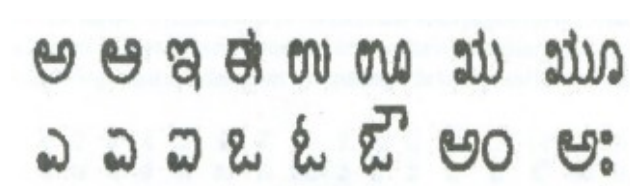

Table 2. Consonant – Vowel combination

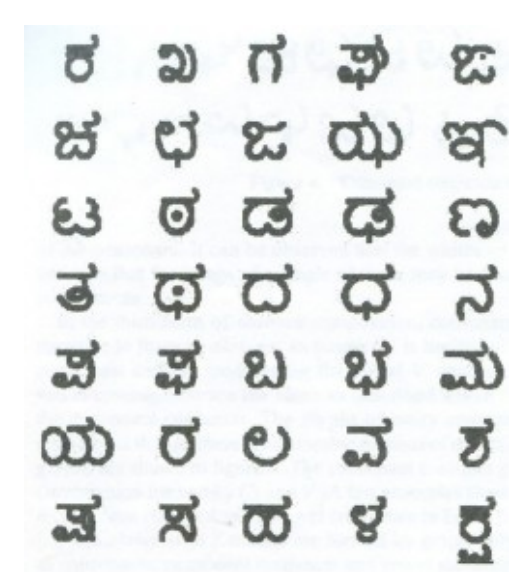

Table 3. Consonant Conjunct

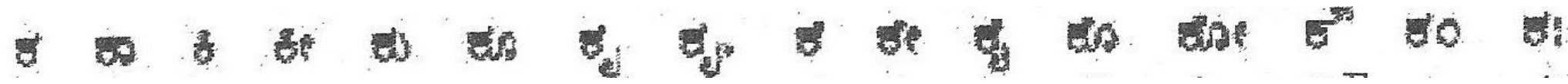

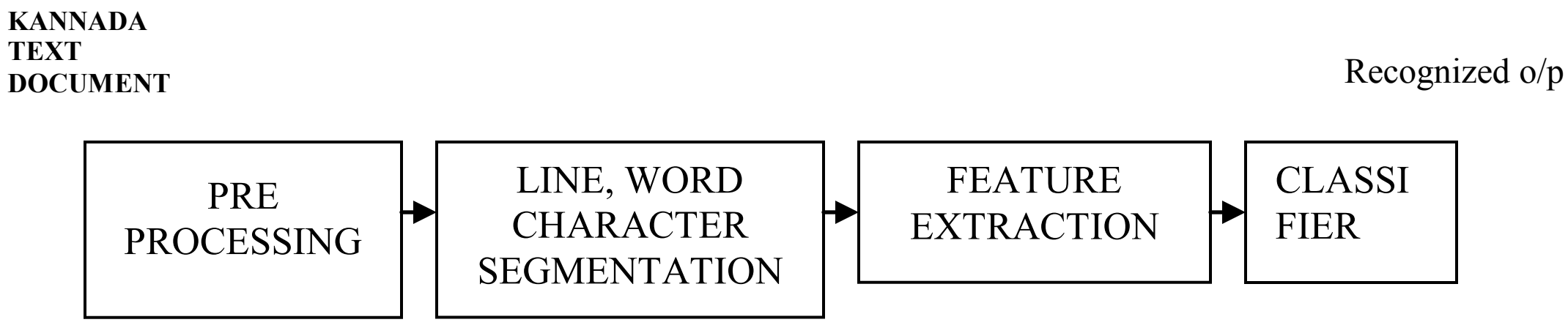

Figure 1. Block Diagram of a KCR system

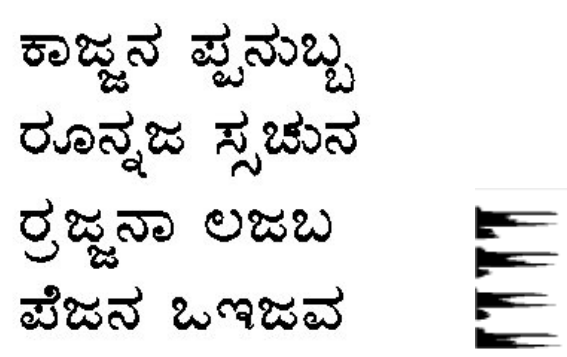

Figure 2. Text lines with HPP

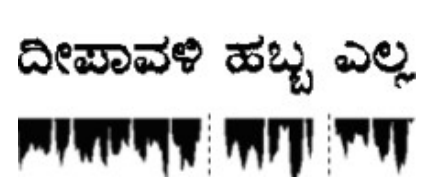

Figure 3. Text lines with VPP

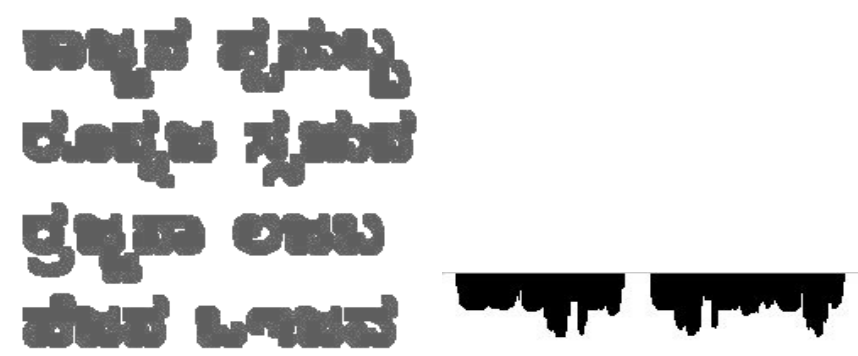

Figure 4. Dilated image and VPP of line 1

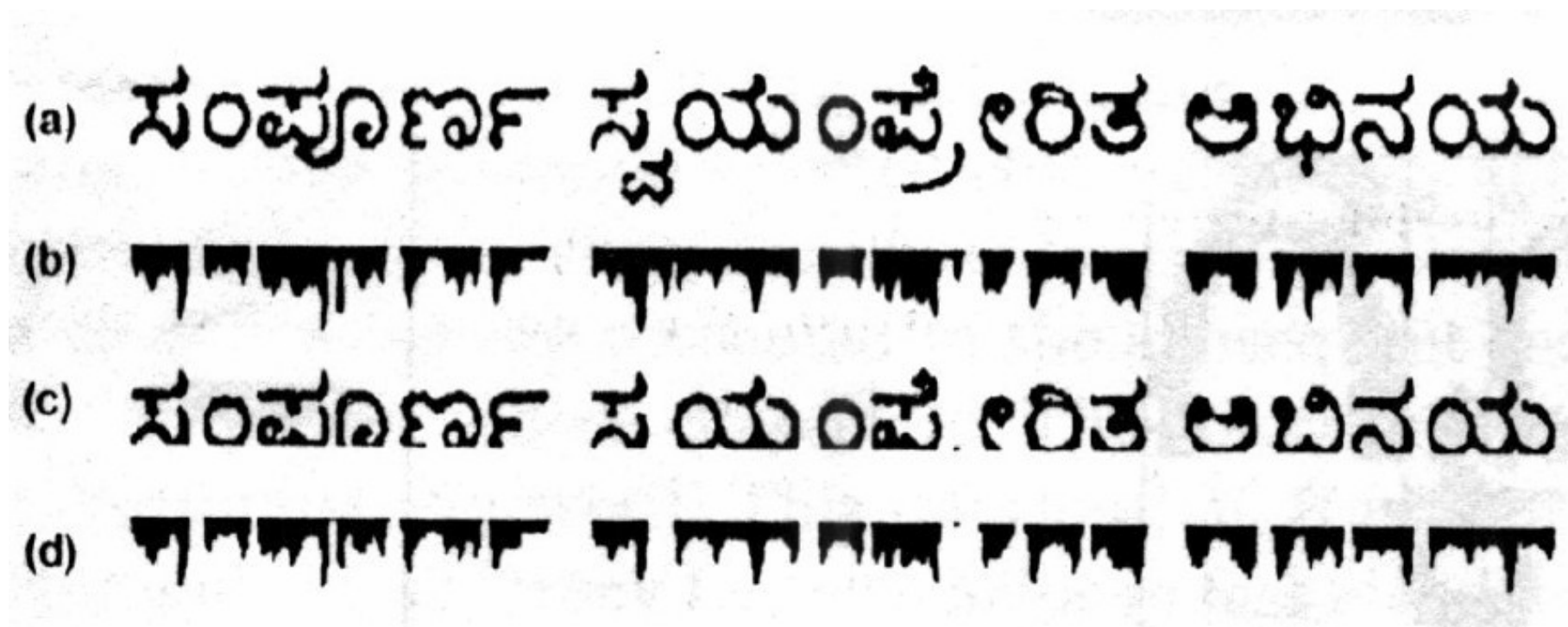

Figure 5 **(a)** Text line (**b**) VPP **(c)** Middle and top zone **(d)** VPP of (c)

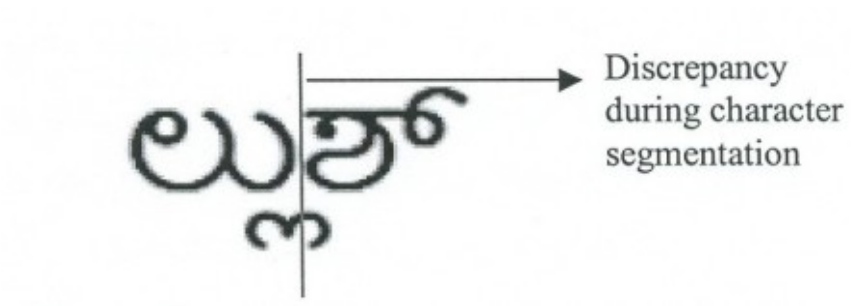

Figure 6**.** Discrepancy in Character Segmentation

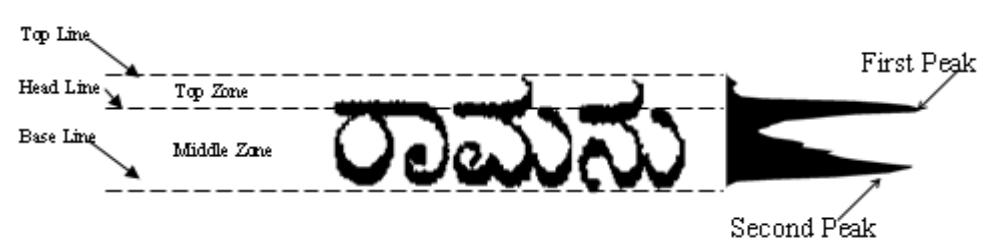

Figure 7. Two horizontal zones in a word without subscripts

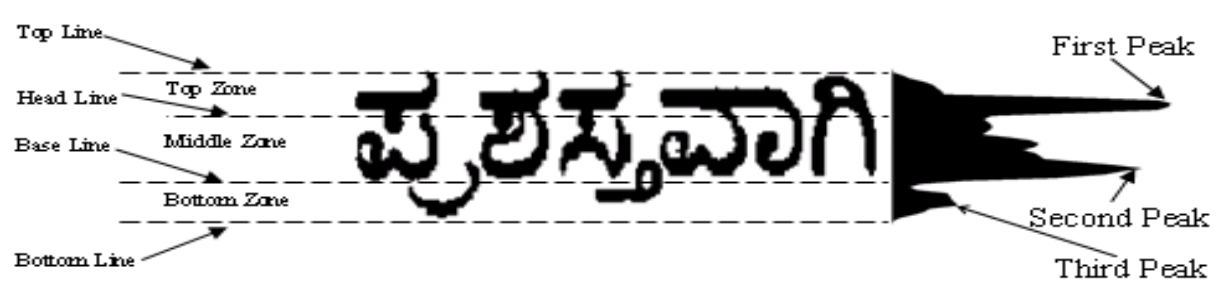

Figure 8**.** Three horizontal zones in a word with subscripts

Table 4. Different Feature Extraction Techniques

| **Sl No** | **Publication** | **Feature Extraction** |
|---|---|---|
| 1. | Kumar and Ramakrishnan [8] | DCT, DWT ( Haar, Db2), KLT. |
| 2. | Ashwin and Sastry [1] | Division of segments into tracks and sectors. |
| 3. | Kunte and Samuel [11] | Hu's invariant moments, Zernike moments. |
| 4. | Kunte and Samuel [11] | Fourier and Wavelet Descriptors |
| 5. | Kunte and Samuel [11] | Contour extraction and wavelet transform. |
| 6. | Kumar and Ramakrishnan [6] | Subspace projection. |

Table 5: Nearest Neighbor classifier on various features

| Sl. No | Publi cation | Symbol | Features | Feature dimension | % Recognition Rate | Recognition time(min) |
|---|---|---|---|---|---|---|
| 1 | [8] | Base character | DCT | 16 (4 x 4) | 93.80 | 0.60 |
| | | | | 64 (8 x 8) | 98.70 | 1.23 |
| | | | | 144 (12 x 12) | 98.27 | 2.11 |
| | | | KLT | 40 | 98.70 | 0.92 |
| | | | | 50 | 98.55 | 1.13 |
| | | | | 60 | 98.77 | 1.32 |
| | | | DWT (Haar) | 64 (8 x 8) | 98.83 | 2.65 |
| | | | DWT (db2) | 100 (10 x 10) | 98.55 | 3.53 |
| 2 | [1] | Base character | Zernike | 7 | 92.66 | - |
| | | | Structural (equally spaced radial Tracks & sectors) | 48 | 92.76 | - |
| | | | Modified structural (adaptively spaced for Equal ON Pixels in each annular region) | 48 | 93.28 | |
| 3 | [8] | Top and right matra | Haar (db1) | 64 | 94.20 | 0.56 |
| | | | DCT | 64 | 93.04 | 0.43 |
| | | Consonant conjuncts | Haar (db1) | 64 | 96.61 | 0.37 |
| | | | DCT | 64 | 96.7 | 0.21 |
| | | | Zern like | 7 | 88.13 | - |
| 4 | [1] | Top matra | Structural | 48 | 86.91 | - |
| | | | Modified | 48 | 88.28 | |
| | | Consonant | Zernike | 7 | 92.22 | |
| | | | Structural | 48 | 89.27 | |
| | | | Modified | 48 | 92.76 | |
| 5 | [6] | Base | Subspace | 60 | 94.5 | 1.69 |

Table 6. Neural Network classifier on various features

| Sl. No | Publication | Symbol | Features | Feature Dimension | No of Hidden Neurons | Spread of Gaus sian | Type of Neural Network | % Recognition rate | Type of neural network | % Recognition rate |
|---|---|---|---|---|---|---|---|---|---|---|
| 1 | [6] | Base character | DWT | 64 | 200 | 10 | RBF | 96 | **RBF with structural features** | 97.5 |
| | | | | 64 | 1200 | 10 | RBF | 97.3 | RBF with structural features | 99.1 |
| 2 | [8] | Base character | Haar | 16 | 20 | 4 | BPN | 96.1 | RBF | 69.2 |
| | | | | | 25 | 11 | | 96.3 | | 99.0 |
| | | | DCT | 16 | 20 | 4 | | 95.9 | | 52.9 |
| | | | | | 25 | 11 | | 95.7 | | 98.9 |
| 3 | [1] | | Zernike | **7** | | | SVM | 92.6 | | |
| | | | Structural | 48 | | | | 93.8 | | |
| | | | Modified structural | 48 | | | | 93.3 | | |
| 4 | [8] | Top and Right Matras | Haar | 64 | - | 10 - | **-** | - | RBF | 96.8 |
| | | | DCT | 64 | -- | | **---** | --- | | 96.8 |
| | | Top matra | Zernike | 7 | | | SVM | 88.4 | | – |
| | | | Structural | 48 | | | | 88.0 | | |
| | | | Modified structural | 48 | | | | 87.2 | | |
| 5 | [8] | Consonant Conjunct | Haar | 64 | 50 | 10 | BPN | 95.1 | RBF | 95.7 |
| | | | DCT | 64 | 50 | 10 | | 93.8 | | 95.5 |
| | | | Zernike | 7 | | | SVM | 91.9 | | |
| | | | Structural | 48 | | | | 93.8 | ——— | |
| | | | Modified Structural | 48 | | | | 94.9 | | |
| 6 | [11] | Base Character | Hu's moments | 7 | Equal to number of training samples | | | | RBF | 82 |
| | | | Zernike moments | 7 | | | | | RBF | 91.8 |
| | | | | 10 | | | | | | 96.8 |
| 7 | [14] | Base Character | DWT | 120 | 60 | | BPN | 92 | --- | ---- |
| | | Consonant | | | | | | | | |

## BIOGRAPHY

**Dr. S. Sethu Selvi** is a Professor and Head of department of Electronics and Communication Engineering, MSRIT, Bangalore. She has published 15 papers in International, National journals and conference proceedings. Her areas of interest are Digital Signal Processing, Digital Image Processing, and Digital Signal Compression.

**K. Indira** is an Assistant Professor in the department of Electronics and Communication Engineering, MSRIT, Bangalore. Pursuing Ph.D. in Image Processing under VTU. She has published 6 papers in National and International Conference Proceedings. Her areas of interests are Digital Signal Processing and Digital Image Processing.